\documentclass[11pt]{article}

\usepackage[final]{acl}

\usepackage{times}
\usepackage{latexsym}

\usepackage[T1]{fontenc}

\usepackage[utf8]{inputenc}

\usepackage{microtype}

\usepackage{inconsolata}

\usepackage{graphicx}
\usepackage{microtype}
\usepackage{graphicx}
\usepackage{subfigure}
\usepackage{booktabs} 

\usepackage{amsmath}
\usepackage{amssymb}
\usepackage{mathtools}
\usepackage{amsthm}

\usepackage{array}    
\usepackage{amsmath}
\usepackage{tabularx}
\usepackage{seqsplit} 
\usepackage{ragged2e} 
\usepackage{graphicx}
\usepackage{multirow}
\usepackage{caption} 
\usepackage{cuted}   

\title{Can the capability of  Large Language Models be described by human ability? A Meta Study}

\author{
 \textbf{Mingrui Zan\textsuperscript{1,2,3}},
  \textbf{Yunquan Zhang\textsuperscript{2}},
  \textbf{Boyang Zhang\textsuperscript{1,2,3}},
  \textbf{Fangming Liu\textsuperscript{3}},
\\
\textbf{Daning Cheng\textsuperscript{2}},
\\
  \textsuperscript{1}University of Chinese Academy of Sciences,
  \textsuperscript{2}Institute of Computing Technology,
  \textsuperscript{3}PengCheng Lab,
\\
  \small{
  \textbf{Corresponding Author's Email:} \href{mailto:email@domain}{chengdaning@ict.ac.cn}
  }
}

\begin{document}
\maketitle
 
\begin{abstract}

 Users of Large Language Models (LLMs) often perceive these models as intelligent entities with human-like capabilities. However, the extent to which LLMs' capabilities truly approximate human abilities remains a topic of debate. In this paper, to characterize the capabilities of LLMs in relation to human capabilities, we collected performance data from over 80 models across 37 evaluation benchmarks. The evaluation benchmarks are categorized into 6 primary abilities and 11 sub-abilities in human aspect. Then, we then clustered the performance rankings into several categories and compared these clustering results with classifications based on human ability aspects. Our findings lead to the following conclusions: 1.  We have confirmed that certain capabilities of LLMs with fewer than 10 billion parameters can indeed be described using human ability metrics; 2. While some abilities are considered interrelated in humans, they appear nearly uncorrelated in LLMs; 3. The capabilities possessed by LLMs vary significantly with the parameter scale of the model.
\end{abstract}

\section{Introduction}

In the view of  LLM applications' user, many perceive LLMs as intelligent entities similar to humans, expecting them to possess specific abilities such as analysis and reasoning, akin to human capabilities. Based on these capabilities, we  expect LLMs  to be able to complete various tasks corresponding to their respective abilities. 

However, it remains to be discussed whether the capabilities of LLMs in completing these tasks are similar to human abilities: On one hand, we do not know whether the classification of LLMs' capabilities is analogous to the classification of human abilities. On the other hand, we are also uncertain about the correlation between these abilities in LLMs. For example, when a person excels in mathematics, we naturally assume that he also has strong physics skills. However, it is unknown whether similar correlations exist among the capabilities of LLMs.  What is more, we do not know whether specific capabilities will vary with changes in the scale of the model.

For humans, consistent and stable performance on a type of task is regarded as an ability\cite{zalta1995stanford}. The main way to assess an person's ability. For the difficulty varies across exams, absolute scores should not be used as a basis for comparison. However, within a group (such as a class), rankings should be relatively stable. If multiple exams in the same subject show consistent rankings of student performance, then students with higher rankings are often considered to have higher ability, while those with lower rankings are deemed to have lower ability. In this case, the subject can be regarded as assessing a specific ability, such as math ability. If the rankings in a subject vary widely across exams, it may not be appropriate to consider that subject as assessing a particular ability.

Inspired by the way human being assess student ability, we have applied a similar evaluation approach to large language models. The exams for LLMs are referred to as evaluation. In this paper, we have compiled 37 benchmarks and categorized them into 6 human being's ability measurements and 11 human being's sub-ability measurements based on evaluations' descriptions. Each sub-ability has 2-5 different evaluations, and we have collected results from over 80 models on these benchmarks for analysis.

\begin{figure}[!htbp]
     \centering
     \includegraphics[width=0.4\textwidth]{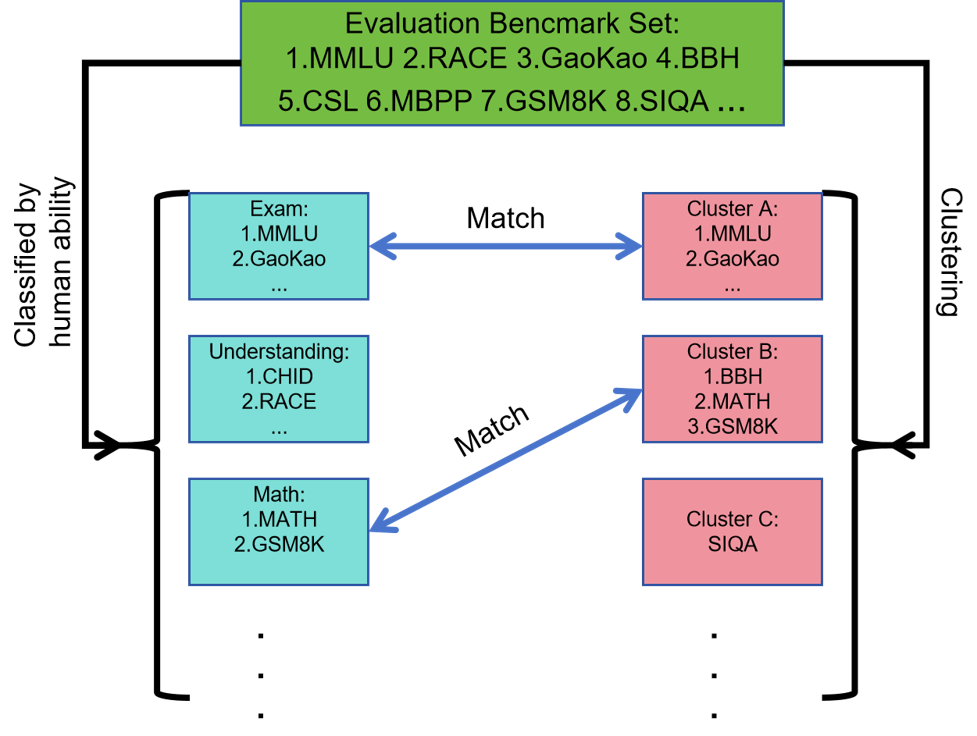}
    \caption{The  Research Frame: By comparing the benchmarks clustering and human ability classification, we can analyze the matching relationship between the capabilities of LLMs and human abilities.}
     \label{frame}
 \end{figure}

We have summarized the rankings of different models on these evaluations and performed clustering based on the ranking distances constructed using the Spearman correlation coefficient to examine the correlations between these evaluations. Finally, we compare and match the clustering results with the classification based on human ability to analyze whether these ability can constitute what is understood as ability in human beings' view. The whole research frame is also show in figure \ref{frame}.

Based on above analysis process, we gain the following conclusions:

1. Our analysis of LLMs with less than 10B parameters shows that their performance in understanding, analysis, examination, text reasoning, mathematical reasoning, knowledge application, and coding is relatively stable. However, their summarization and common-sense reasoning are unstable, for LLMs less than 10B parameters, these latter abilities may not be considered reliable capabilities. 2. By examining similar capabilities in specific tasks—such as analysis and understanding in natural language tasks, and text reasoning and common-sense reasoning in reasoning tasks—we find that similar abilities in human cognition do not necessarily align with those in LLMs, and significant differences may exist. Therefore, when applying LLMs, we cannot assume that the superiority or inferiority of a particular ability will directly correspond to how it is understood in human terms. 3. By comparing the capabilities of 10B models and 10-20B models, we have observed that the emergence and disappearance of model capabilities vary with the changes in model scale.

\section{Related Work}

The evaluation of artificial intelligence models constitutes a critical component in assessing model performance. The emergence of general language models such as BERT \cite{devlin2018bert} has underscored the growing importance of constructing more comprehensive benchmark tests to evaluate the broad capabilities of these LMs. Popular evaluation benchmarks like GLUE \cite{wang2018glue} and its successor SuperGLUE \cite{wang2019superglue} have been instrumental in evaluating the performance of language models across a range of natural language processing (NLP) tasks, including semantic similarity assessment, sentiment analysis, among others. The GLUE suite of evaluation benchmarks has had a profound impact on the development of language models, spurring researchers to enhance the generalization performance of their models.

As large language models increasingly attract widespread attention, the development of their evaluation methods is also accelerating. In recent years, numerous studies have been dedicated to evaluating large language models from multiple dimensions. For instance, in 2016, Baotian Hu\cite{hu2015lcsts}  constructed a large-scale Chinese short text summarization dataset and used the ROUGE metric to assess the similarity between model-generated summaries and human-written summaries, thereby measuring the model's performance. In 2018, Shashi Narayan et al.\cite{narayan2018don}  employed a combination of ROUGE automatic evaluation and human evaluation to score the capabilities of language models in extreme summarization tasks. In 2022, YuDong Li et al. \cite{li2022csl} proposed a evaluation benchmark based on CSL to evaluate the performance of models in tasks such as summarization generation, keyword extraction, and text classification. By 2023, Zorik Gekhman et al.\cite{gekhman2023trueteacher}  used the TrueTeacher method to assess the consistency between model-generated summaries and facts.

Many works try to give summary the evaluation benchmarks \cite{DBLP:journals/tist/ChangWWWYZCYWWYZCYYX24}. However, they fail to answer the questions: Are these evaluation benchmarks are related? Do these benchmarks indicate the model's capabilities is like the human beings abilities? The model's capabilities can be existed all the time?

\section{Clustering method}
\subsection{Spearman's rank correlation coefficient}

Spearman's rank correlation coefficient is a nonparametric measure of the monotonicity of the relationship between two variables\cite{spearman2010proof}. It is suitable for analyzing the correlation of nonlinear and non-normally distributed variables, using a monotonic function to evaluate the correlation between two statistical variables. Because Spearman's rank correlation coefficient is calculated based on the ranking order of data points rather than their exact values, it is robust against outliers and will not be significantly skewed by extreme values. 

For a sample of size $n$, the $n$ original data pairs $X_i$, $Y_i$ are converted into rank order data $R(X_i)$, $R(Y_i)$, and the correlation coefficient rs is expressed as follows:
\begin{equation}
      r_{s}=\rho[R[X],R[Y]]=\frac{\operatorname{cov}[R[X],R[Y]]}{\sigma_{R[X]}\sigma_{R[Y]}}\label{s}
\end{equation}
Where $\rho$ is the Pearson correlation coefficient, but calculated using rank variables, ${cov}[R[X],R[Y]]$ is the covariance of the rank variables, and $\sigma_{R[\cdot]} $ are the standard deviations of the rank variables.

The range of the Spearman's correlation coefficient is from [-1,1]. When $\rho$=1, it indicates perfect positive correlation; as one variable increases, the other also increases in the same order. When $\rho$=-1, it indicates perfect negative correlation; as one variable increases, the other decreases in the opposite order. When $\rho$=0, it means there is no monotonic relationship between the two variables.

In practical applications, there is no strictly defined "threshold" to determine whether the correlation is significant or important. Generally, $|r_{s}| < 0.3$  indicates weak or almost non-existent correlation. When the Spearman's coefficient is close to zero, it suggests that the monotonic relationship between the two variables is very weak or does not exist; $0.3 \leq |r_{s}| < 0.7$. indicates a moderate degree of correlation between the two variables; $|r_{s}| \geq 0.7$. indicates a strong correlation between the two variables.
\subsection{Cluster based on Spearman's coefficient}
Given the variability in capabilities among models, there are significant differences in score ranges across different evaluation systems for LLMs. For instance, among the 52 LLMs with parameter sizes not exceeding 7 billion included in this study. The highest score in the OpenBookQA\cite{mihaylov2018can} was 92.2, while the lowest was 22.4. In contrast, under the Math\cite{hendrycks2021measuring}, the highest score was 19.2 and the lowest was 0.6. If absolute scores are used as the sole criterion for classification, it fails to accurately reflect the relative performance of different models within the same evaluation system, and may also lead to the neglect of certain features in evaluation systems where the models' capabilities are weaker (with lower score ranges).

Therefore, this paper adopts the Spearman's coefficient as the basis to construct distances between variables for cluster analysis. This method is independent of the specific numerical values of the variables and focuses on their relative ranks. By this approach, the similarities and differences between different evaluations can be effectively revealed. 

Based on the properties of Spearman's rank correlation coefficient Eq.  \eqref{s}, this study performs  clustering of different variables using $d_{s}$.
\begin{equation}
      d_{s}=\sqrt{2(1-r_{s})}
\end{equation}
This equation sets the distance between perfectly positively correlated variables to 0, the distance between uncorrelated variables to approximately \(\sqrt{2}\), and the distance between perfectly negatively correlated variables to 2.  Clearly, as $r_{s}$ decreases, it indicates a weaker monotonic relationship between the two variables, resulting in a larger $d_{s}$. Conversely, as $r_{s}$ increases, it indicates a stronger monotonic relationship between the two variables, leading to a smaller $d_{s}$. We utilize hierarchical clustering to categorize the evaluation benchmarks, considering  that, compared with other clustering methods like K-means, the predetermined number of clusters is not feasible in this research: we have to determine  number of clusters by Spearman's rank correlation coefficient. What is more, the evaluation benchmarks data forms non-convex clusters.

\section{Evaluation Benchmark Analysis}
According to the Scaling Laws proposed by Kaplan et al\cite{kaplan2020scaling}, the scale of language model parameters affects the performance of downstream tasks. If the performance of models of different sizes on the same evaluation benchmark is statistically analyzed, the main factors influencing rankings will shift from the models' inherent capabilities to their scale. Therefore, this chapter only considers the scoring of LLMs with less than 10B parameters on various evaluations.
\subsection{Evaluation Benchmarks Set}
Some simple evaluation benchmarks consist of a single task on a single dataset, which is also the common basic evaluation pattern for natural language processing. To comprehensively evaluate large language models, more general evaluation benchmarks integrate and reorganize multiple datasets. This section focuses on the evaluation of large language models, organizing, summarizing, and classifying the current evaluation paradigms. 

\autoref{tab:benchmarks} lists the evaluation benchmark and their capability classification (categorized by human abilities). These classifications are based on the survey works \cite{DBLP:journals/tist/ChangWWWYZCYWWYZCYYX24, minaee2024large} and  descriptions of their capabilities provided in the literature of each benchmark, more detail is shown in appendix. The existing ability tasks can be divided into six categories: natural language, examination, reasoning, knowledge, multilingual, and coding.
\renewcommand{\arraystretch}{1.2}

\begin{table*}[ht]
\centering
\small 
\begin{tabularx}{\textwidth}{ X >{\raggedright\arraybackslash}X X >{\raggedright\arraybackslash}X X >{\raggedright\arraybackslash}X }
\toprule
\textbf{Evaluation} & \textbf{Classification} & \textbf{Evaluation} & \textbf{Classification} & \textbf{Evaluation} & \textbf{Classification} \\
\midrule
C$^3$\cite{sun2020investigating} & \multirow{6}{=}{Understanding} & AGIEval\cite{zhong2023agieval} & \multirow{6}{=}{Examination} & AFQMC\cite{zhang2022fengshenbang} & \multirow{4}{=}{Text reasoning} \\
CHID\cite{zheng2019chid} &  & ARC-C\cite{clark2018think} &  & CMNLI\cite{xu2020clue} &  \\
Drop\cite{dua2019drop} &  & C-Eval\cite{huang2024c} &  & COPA\cite{wang2019superglue} &  \\
OpenBookQA\cite{mihaylov2018can} &  & CMMLU\cite{li2023cmmlu} &  & OCNLI\cite{xu2020clue} &  \\\cline{5-6}
RACE\cite{lai2017race} & & GAOKAO-bench\cite{zhang2023evaluating} &  & HellaSwag\cite{zellers2019hellaswag} & \multirow{3}{=}{Commonsense reasoning}  \\
ReCoRd\cite{wang2019superglue} &  & MMLU\cite{hendrycks2020measuring} &  & PIQA\cite{bisk2020piqa} &   \\\cline{1-2}\cline{3-4}
AX$_g$\cite{wang2019superglue} & \multirow{4}{=}{Analysis} & BoolQ\cite{clark2019boolq} & \multirow{4}{=}{Knowledge} & SIQA\cite{sap2019socialiqa} &  \\\cline{5-6}
LAMBADA\cite{paperno2016lambada} &  & Common-senseQA\cite{talmor2018commonsenseqa} &  & GSM8K\cite{cobbe2021training} & \multirow{2}{=}{Mathematical reasoning} \\
WIC\cite{pilehvar2018wic} &  & NQ\cite{kwiatkowski2019natural} &  & MATH\cite{hendrycks2021measuring} &  \\\cline{5-6}
WSC\cite{levesque2012winograd} &  & TriviaqQA\cite{joshi2017triviaqa} & & BBH\cite{suzgun2022challenging} & Comprehensive Reasoning \\\cline{1-2}\cline{3-6}
CSL\cite{li2022csl} & \multirow{3}{=}{Summarization} & HumanEval\cite{chen2021evaluating} & \multirow{2}{=}{Coding} & Flores\cite{goyal2022flores} & \multirow{2}{=}{Multilingual} \\
LCSTS\cite{hu2015lcsts} & & MBPP\cite{austin2021program} & & TyDiQA\cite{clark2020tydi} &  \\
XSum\cite{narayan2018don} & & & & & \\
\bottomrule
\end{tabularx}
\caption{Evaluation benchmark and  Their Classification}
\label{tab:benchmarks}
\end{table*}

\subsection{Evaluation Benchmark Clustering}

\begin{figure*}[htbp]
    \centering
    \includegraphics[width=0.9\textwidth]{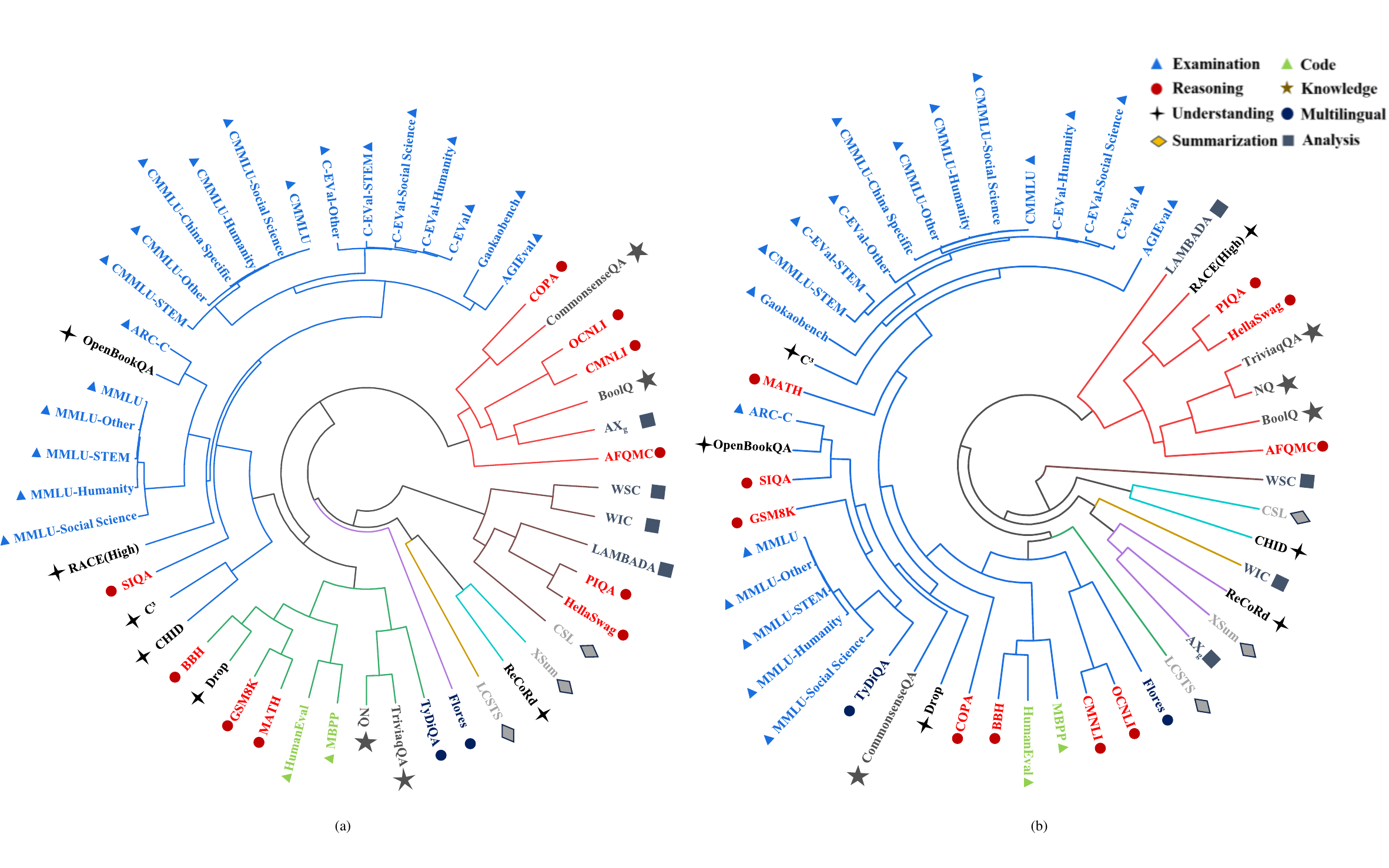} %
    \caption{Evaluation benchmark clustering results for language models with different parameter sizes(a)less than 10B (b)10-20B}
    \label{7clust}
\end{figure*}

We compiled the scores of 52 LLMs (less than 10B) on each evaluation benchmark. It is noteworthy that single linkage, complete linkage, average linkage, and Ward's method in hierarchical clustering do not produce fundamentally different outcomes; therefore, the results presented in this paper are based on the average-linkage agglomerative method. After the cluster analysis, the classification obtained is shown in Figure \ref{7clust} (a). 

To counteract the impact of noise and outliers, which is the mainly defect for  hierarchical clustering, following the clustering process, we will also check the Spearman's  correlation coefficient within each cluster, while comparing them with human-classified outcomes. Figures \ref{fig:figures} display correlation coefficient matrices for different categories based on human abilities. These matrices also can be used to evaluate the strength of the correlation between specific  evaluation benchmarks.

\subsection{Analysis Results}
In this part, we match clustering results with human classification results, and further examine the details of the evaluation benchmarks' Spearman's correlation coefficients within the same cluster.

\subsubsection{Natural language task}

\textbf{Understanding} The intention of developing language models, especially large-scale language models, is to enhance the efficiency of natural language processing tasks. As a key branch of natural language processing, the improvement of natural language understanding is particularly crucial. From the perspective of meeting human needs, it is of vital importance whether LLMs can more accurately understand the tasks of input sequences.

Six evaluation benchmarks for understanding are involved, including Chinese evaluation benchmarks, C³, CHID and English evaluation benchmarks, DROP, OpenBookQA, ReCoRD and RACE.

As can be seen from Figure \ref{7clust} (a)  , most of the evaluation sets related to understanding are classified into one cluster. Figure \ref{fig:understanding} lists the Spearman's coefficients between each evaluation set. It can be seen that the Spearman's coefficient between OpenBook QA and RACE is as high as 0.88, thus they can almost be considered as one evaluation.

Thus, we can conclude that LLMs exhibit a consistent trend in performance on most understanding ability evaluation. Understanding--this human ability can be used as a measure of the capabilities of LLMs.

\textbf{Summarization} In practical application contexts, the task of generating text summaries using language models is an important function within the subfield of Natural Language Generation under NLP. This study conducts an analysis of three evaluation benchmark datasets specifically designed to evaluate the summarization capabilities of LLMs.

We use CSL, Xsum and LCSTS as evaluation benchmarks. As illustrated in Figure \ref{7clust} (a), the three evaluation benchmarks related to summarization capabilities are categorized into three distinct clusters. The detail information is shown in Figure \ref{fig:summary}. This reveals significant variations in the performance of LLMs across different summarization evaluation datasets. A language model may excel in one summarization task but perform inadequately in others. Consequently, Summarization, this human ability may not accurately represent a specific capability of LLMs.


\textbf{Analysis}
Content analysis constitutes a significant capability of LLMs. At present, a substantial number of evaluations pertain to analysis tasks. In this study, we selected the evaluation benchmarks AX$_g$, LAMBADA, WSC, and WIC  for our analysis.


As illustrated in Figure \ref{7clust} (a) and Figure \ref{fig:analysis}, the majority of evaluation benchmarks associated with analytical capabilities are classified within a single category. LLMs  show consistent performance. Therefore, analysis ability can be considered an indicator of model capability.

\textbf{Sub-ability Summary}
Notably, for humans, the abilities to understand, analyze, and summarize a context are closely related; an individual who can effectively understand a text is often capable of performing robust analysis and summarization. In contrast, for Large Language Models , these three capabilities—understanding, analysis, and summarization are categorized separately(Figure \ref{7clust} (a)) and are considered distinct capabilities.

\subsubsection{Examination Task}
The capacity of Large Language Models to aid students and professionals in exam preparation and success is a crucial metric for evaluating their performance. Presently, a significant number of  evaluation benchmarks focus on this capability. Several evaluation benchmarks based on subject-specific exam content have been developed.

In this paper, we use AGIEval, CMMLU, C-Eval, GAOKAO-bench, MMLU, ARC as evaluation benchmarks. These evaluations are common used evaluation benchmarks, some of them like GAOKAO-bench are gained from the real human being examination, for example, GAOKAO-bench is an intuitive evaluation benchmark that uses questions from China's National College Entrance Examination (GAOKAO) as test samples.

\begin{figure*}[!th]
    \centering

    \begin{minipage}[b]{0.325\textwidth}
        \centering
        \includegraphics[width=\textwidth]{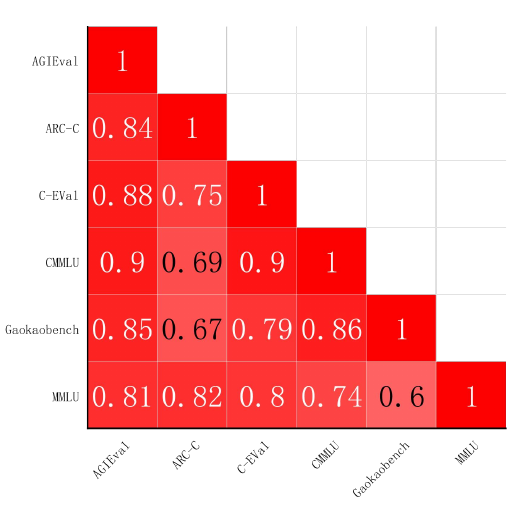}
        \caption{Exam}
        \label{fig:exam}
    \end{minipage}\hspace{0.5em}
    \begin{minipage}[b]{0.325\textwidth}
        \centering
        \includegraphics[width=\textwidth]{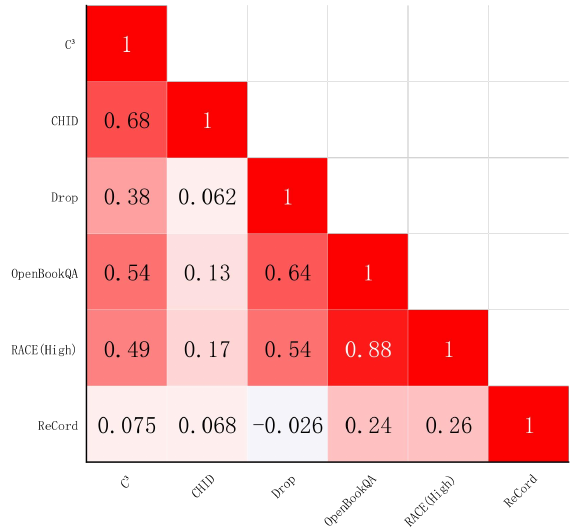}
        \caption{Understanding}
        \label{fig:understanding}
    \end{minipage}\hspace{0.5em}
    \begin{minipage}[b]{0.325\textwidth}
        \centering
        \includegraphics[width=\textwidth]{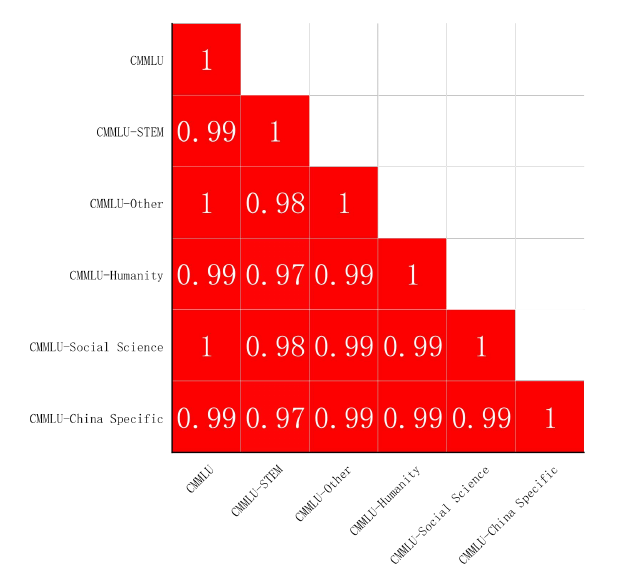}
        \caption{CMMLU}
        \label{fig:cmmlu}
    \end{minipage}\\[\baselineskip]

    \vspace{0.5em}
    \begin{minipage}[b]{0.32\textwidth}
        \centering
        \includegraphics[width=\textwidth]{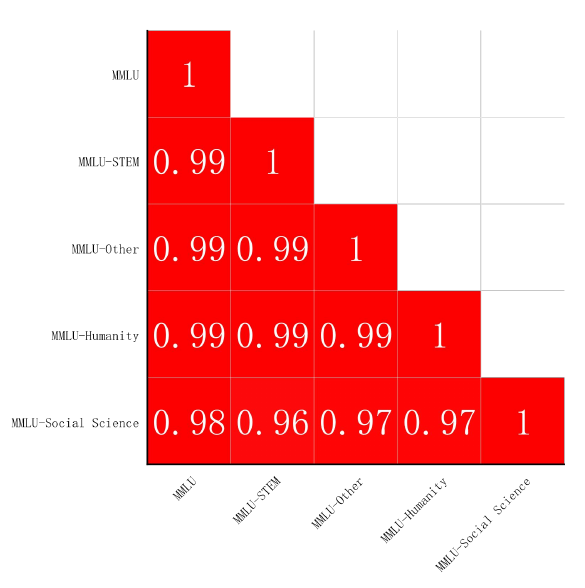}
        \caption{MMLU}
        \label{fig:mmlu}
    \end{minipage}\hspace{0.5em}
    \begin{minipage}[b]{0.32\textwidth}
        \centering
        \includegraphics[width=\textwidth]{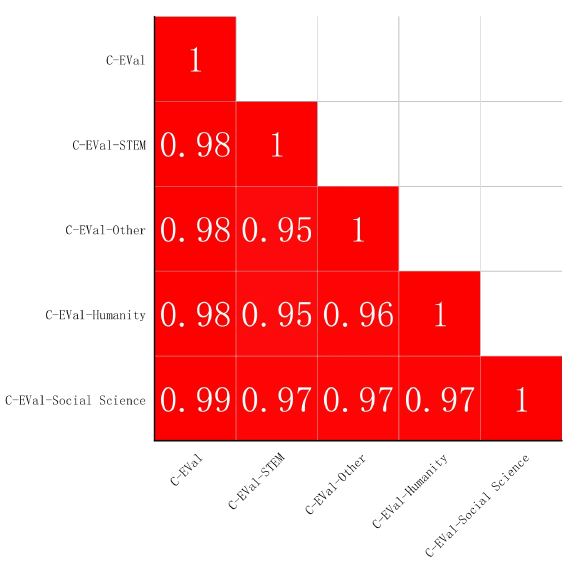}
        \caption{C-Eval}
        \label{fig:c-eval}
    \end{minipage}\hspace{0.5em}
    \begin{minipage}[b]{0.32\textwidth}
        \centering
        \includegraphics[width=\textwidth]{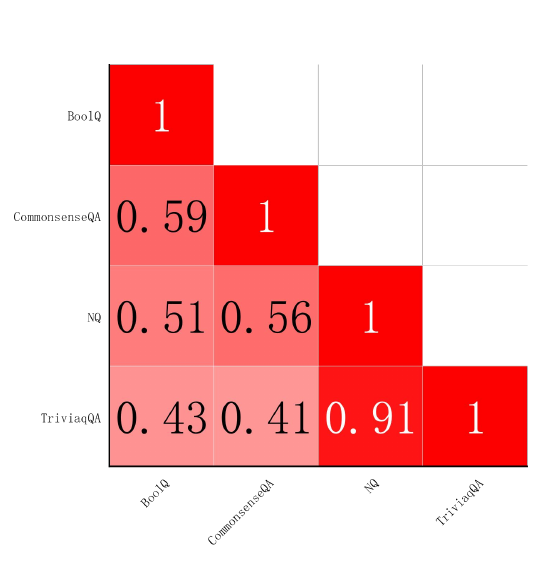}
        \caption{Knowledge}
        \label{fig:knowledge}
    \end{minipage}\hspace{0.5em}
\\[\baselineskip]

    \vspace{0.5em}
    \begin{minipage}[b]{0.191\textwidth}
        \centering
        \includegraphics[width=\textwidth]{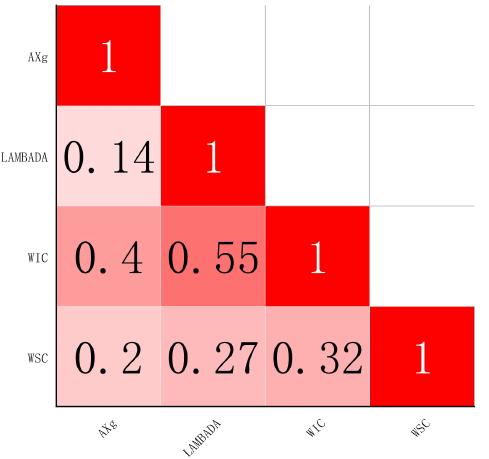}
        \caption{Analysis Ability}
        \label{fig:analysis}
    \end{minipage}
    \begin{minipage}[b]{0.191\textwidth}
        \centering
        \includegraphics[width=\textwidth]{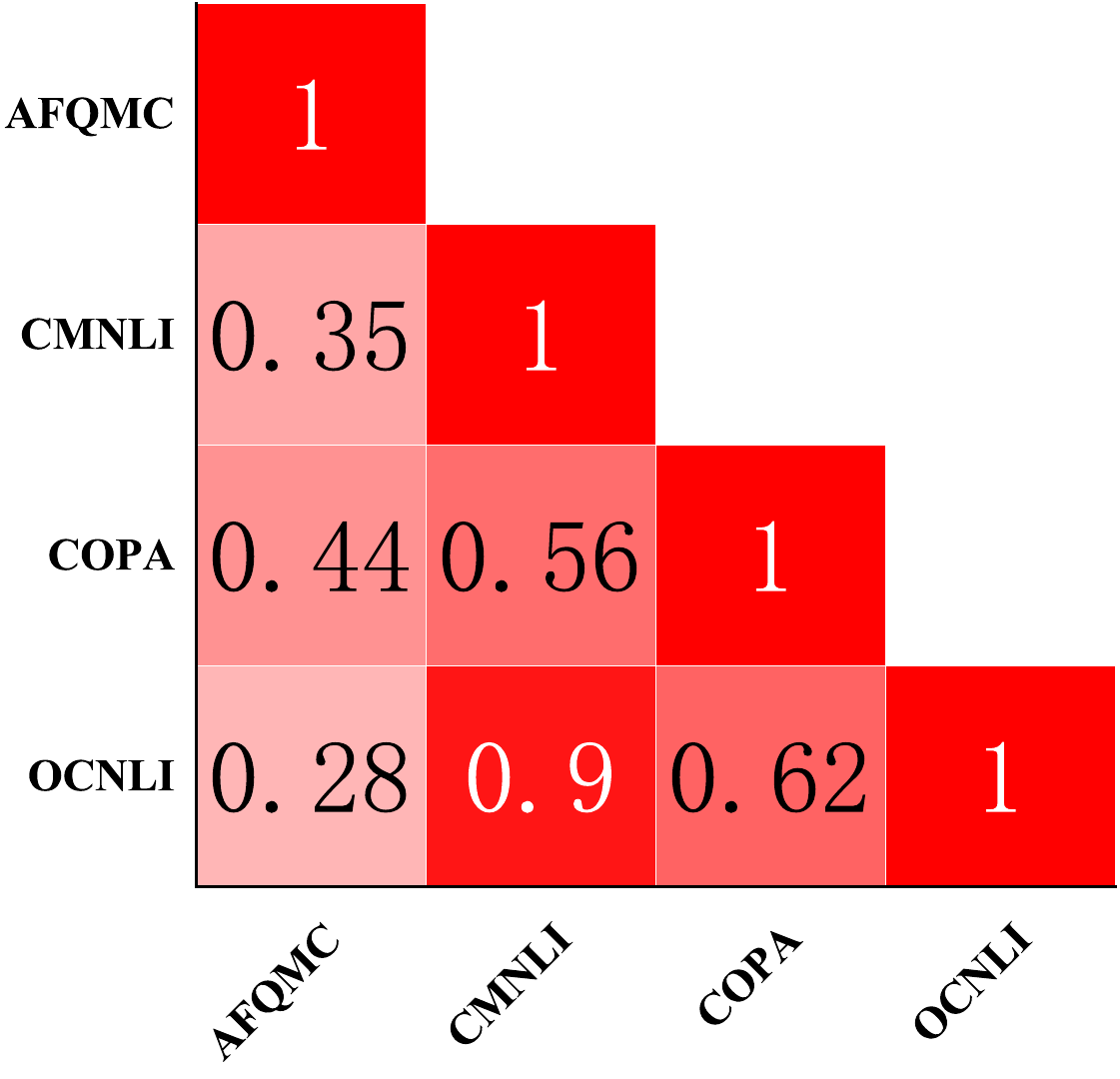}
        \caption{Text-reasoning}
        \label{fig:text}
    \end{minipage}\hspace{0.5em}
    \begin{minipage}[b]{0.191\textwidth}
        \centering
        \includegraphics[width=\textwidth]{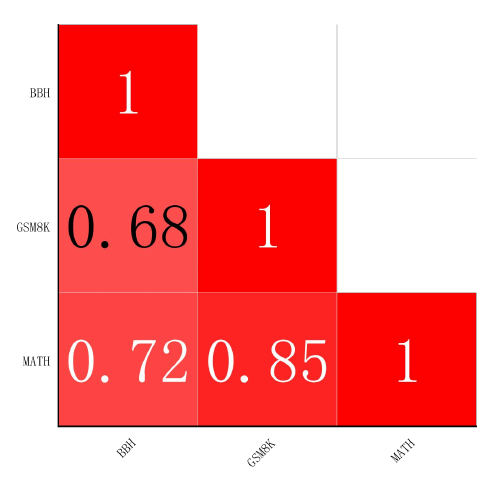}
        \caption{Math-reasoning}
        \label{fig:math}
    \end{minipage}\hspace{0.5em}
    \begin{minipage}[b]{0.191\textwidth}
        \centering
        \includegraphics[width=\textwidth]{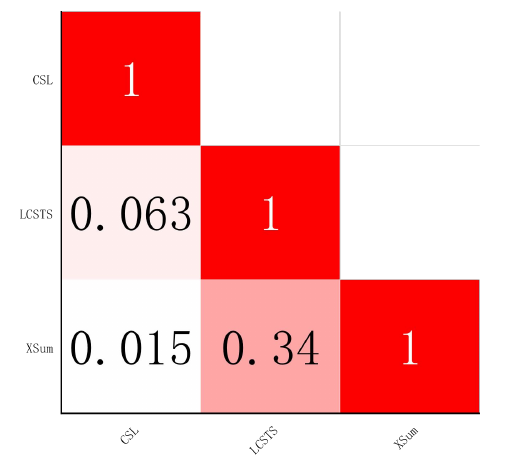}
        \caption{Summarization}
        \label{fig:summary}
    \end{minipage}\hspace{0.5em}
    \begin{minipage}[b]{0.191\textwidth}
        \centering
        \includegraphics[width=\textwidth]{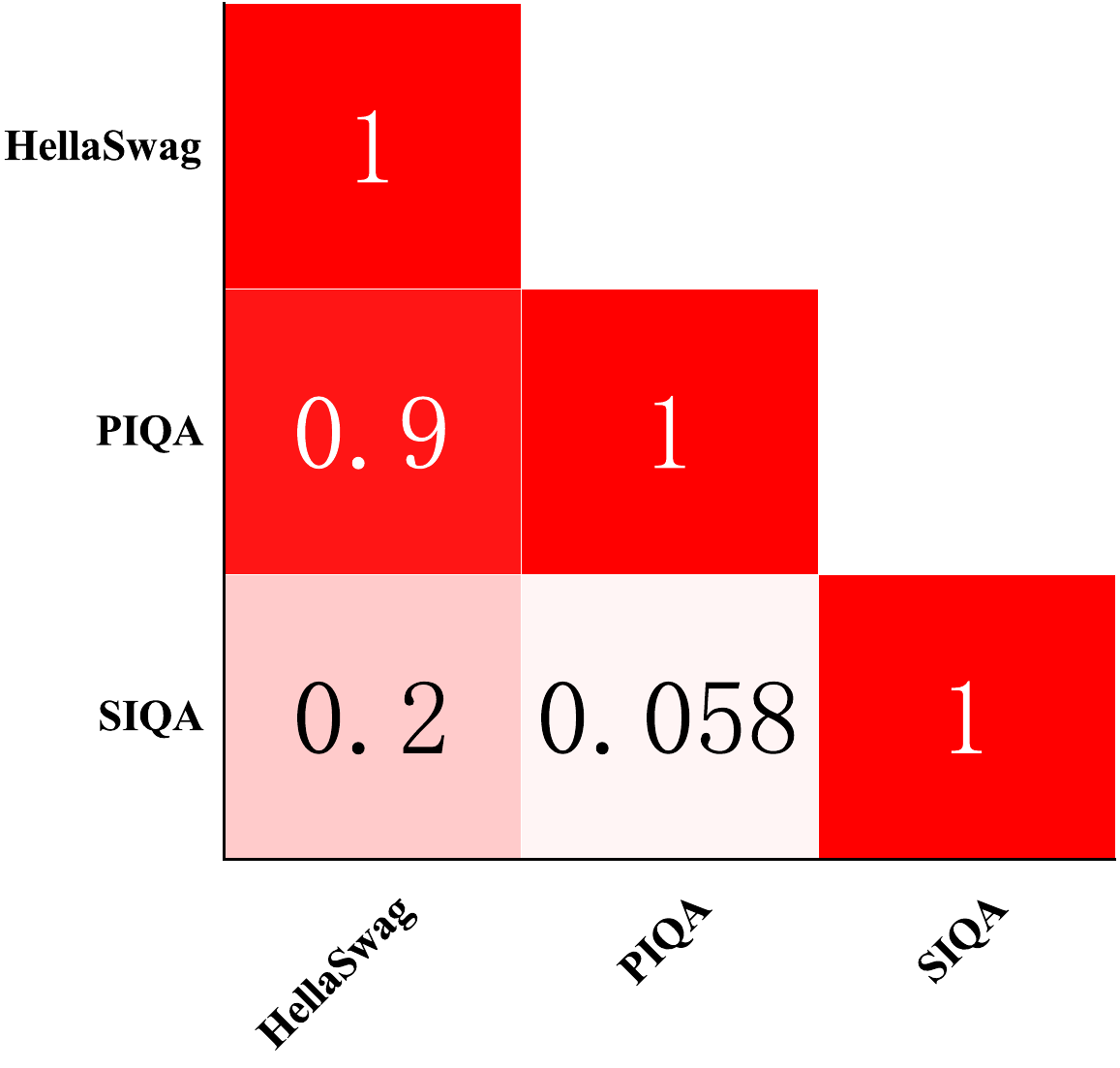}
        \caption{Commonsense}
        \label{fig:commonsense}
    \end{minipage}
   \caption{The Spearman's correlation coefficient matrices in different tasks.  White font represents highly  relevant. The redder the background, the more relevant it is.}
    \label{fig:figures}
\end{figure*}

As shown in Figure \ref{7clust} (a), the six examination-related evaluation benchmarks mentioned above are all categorized into a single cluster, exhibiting positive correlations. Moreover, the Spearman's coefficients between the five evaluation benchmarks excluding GAOKAO-bench are all greater than 0.7 (Figure \ref{fig:exam}). This suggests that as the volume of subject-specific exam data in the evaluation benchmarks increases, the performance of Large Language Models (LLMs) on these evaluation benchmarks tends to stabilize.

Additionally, it is noteworthy that in tests with subcategories such as C-EVAL (Figure \ref{fig:c-eval}), MMLU (Figure \ref{fig:mmlu}), and CMMLU (Figure \ref{fig:cmmlu}), the Spearman coefficients between the scores of individual subcategory tests (e.g., STEM, humanities, social sciences, etc.) and the overall ranking are very high. This indicates that the rankings of LLMs across these evaluations are highly similar.

The above findings suggest that the performance of Large Language Models in examination related evaluations exhibits consistency. Consequently, examination ability can be regarded as a recognizable and distinct capability of LLMs.

\subsubsection{Reasoning Task}
In the application of LLMs by users, reasoning ability plays an evidently critical role. Nevertheless, reasoning ability is not a singular concept but rather encompasses various dimensions, including text reasoning, commonsense reasoning, and mathematical reasoning.

\textbf{Text Reasoning}
Regarding the benchmarks associated with text reasoning tasks, it was noted that AFQMC exhibits relatively lower Spearman's correlation coefficients when compared to the other three benchmarks. However, COPA, OCNLI, and CMNLI show strong correlations among themselves(Figure \ref{fig:text}).

Overall, the performance of LLMs on text reasoning tasks shows a significant level of uniformity. This suggests that despite variations in specific benchmarks or datasets, LLMs tend to exhibit similar capabilities across different types of text reasoning challenges. Such consistency may reflect shared underlying mechanisms or common approaches in how these models process and understand textual information for reasoning purposes.

\textbf{Commonsense Reasoning}
Regarding common sense reasoning tasks, it can be observed that HellaSwag and PIQA are categorized together, whereas SIQA is notably distant from these two benchmarks (Figure \ref{7clust} (a)). The Spearman's correlation coefficients between SIQA and the other two benchmarks, HellaSwag and PIQA, are as low as 0.2 and 0.058(Figure \ref{fig:commonsense}). 

These findings highlight that LLMs do not exhibit uniform performance across different common sense reasoning tasks. Specifically, while HellaSwag and PIQA may require comparable reasoning capabilities, SIQA appears to evaluate a different set of capabilities or challenges. 

\textbf{Mathematical and Comprehensive Reasoning}
As shown in Figure \ref{7clust} (a), MATH and GSM8K exhibit a positive correlation, with a Spearman coefficient as high as 0.855 (see Figure \ref{fig:math} ). This indicates that LLMs show consistent performance in mathematical abilities. Furthermore, the BIG-Bench Hard (BBH), which evaluates comprehensive reasoning capabilities, also shows a correlation with mathematical reasoning.

\textbf{Sub-ability Summary}
The tasks in the above reasoning-related evaluation benchmarks all require one or more steps of inference. LLMs exhibit consistent performance within each reasoning-related evaluation benchmark, but their performance varies significantly across different categories of reasoning evaluation benchmarks. Therefore, when analyzing the capabilities of LLMs, reasoning ability should not be considered a monolithic capability but should be examined separately for each type.

\subsubsection{Knowledge Application Task}
The knowledge application ability of models refers to their capability to effectively leverage the vast amount of knowledge accumulated during training to perform specific tasks or address novel problems. Some evaluation benchmarks have been developed to assess this critical capacity. We use Commonsense QA, NQ, TriviaQA and BoolQ as evalutation benchmarks.

As illustrated in Figure \ref{7clust}(a), the four above evaluation benchmarks do not cluster together. NQ and TriviaQA form one cluster, whereas Commonsense QA and BoolQ constitute a separate cluster. The Spearman's coefficients between evaluation benchmarks from different clusters are notably lower (see Figure \ref{fig:knowledge} ), reflecting reduced correlations. This observation indicates that the performance of Large Language Models (LLMs) in applying knowledge is inconsistent across different evaluation benchmark categories. Therefore, knowledge application cannot be considered a stable capability for evaluating LLMs.

\subsubsection{Multilingual Task}


Translation is an important application scenario for large language models. At present, numerous examples suggest that LLMs can perform well in translation within a single language. However, mutual translation across multiple languages remains one of the challenges for LLMs, which is also reflected in the evaluation and analysis of multilingual capabilities of large language models conducted in this study. This paper, we use Flores and TYDIQA as evaluation benchmarks. 

As shown in Figure 1, the two above evaluation benchmarks do not fall into the same category, and their Spearman's coefficient is 0.26. This indicates that LLMs exhibit significantly different performance across these two evaluation benchmarks.

Therefore, multilingual capabilities should not be evaluated as a single cluster, unified ability for LLMs but should instead be assessed separately for each evaluation benchmark.

\subsubsection{Coding Task}
Auxiliary programming is an important application scenario for LLMs at this stage, exemplified by tools such as GitHub Copilot, Tongyi Lingma, MarsCode and so on. Auxiliary programming plays a significant guiding role for programmers in aspects such as assisted code generation and comment creation.

This paper statistically examines two evaluation benchmarks related to code generation capabilities: HumanEval and MBPP.

As shown in Figure \ref{7clust}(a), not only do the two evaluation benchmarks exhibit a correlation, and their Spearman's coefficient is also as high as 0.78. This suggests that LLMs show consistent performance when addressing code-related problems. Therefore, coding ability can be recognized as a distinct capability of models.

\subsection{Analysis For 10-20B Parameter Model}
We also collected evaluation scores for 30 LLMs with parameter sizes between 10 and  20 billion across 52 evaluation benchmarks. These scores were ranked and treated as a variable for clustering analysis based on the Spearman's  coefficient. The results are shown in Figure \ref{7clust}(b). It is evident that more capabilities are grouped into a single cluster for LLMs with fewer than 10 billion parameters. In this analysis, the largest cluster comprised 24 samples. For LLMs with parameter sizes between 10 and 20 billion, the largest cluster included up to 34 samples. This indicates that as the model parameter size increases, the performance of LLMs across various evaluation benchmarks tends to converge.

Additionally, it was observed that evaluation benchmarks related to knowledge application were clustered together. In the ranking of evaluation scores for models with fewer than 10 billion parameters, these evaluation benchmarks did not cluster together. Furthermore, clusters of samples related to understanding abilities, which were previously grouped together for smaller models, did not remain cohesive when the parameter size increased to 10-20 billion. This suggests that as the model parameter size grows, certain existing capabilities lose stability while new capabilities emerge.

\section{Discussion and Application}
In this paper, we conduct a further exploration of the capabilities of LLMs, delineating the  properties of their capabilities. Our research findings can be applied in different scenarios, including: 1. For evaluation benchmarks grouped under the same capabilities, it suffices to evaluate just one representative, such as different subclasses within CMMLU or the evaluation benchmarks in examination groups, like MMLU and AGIEval. This is because these evaluations exhibit strong correlations, making their ranking results likely to be similar. 2. The capabilities of LLMs do not have an implication relationship among themselves. Therefore, when users have new requirements for LLMs, they should conduct new evaluations rather than speculate based on analogous needs as understood by humans. 3. When assigning a task to a LLM, one should select a model of an appropriate scale. For instance, we cannot expect a model with fewer than 10 billion parameters to handle reasoning tasks effectively.

\section{Conclusion}
In this paper, we collect performance data from over 80 LLMs across 37 benchmarks categorized into 6 abilities and 11 sub-abilities. By clustering these performance rankings and comparing them with human ability classifications, we gained insights into model capabilities. We found that some capabilities of smaller models (less than 10B) can be described using human ability. What is more, certain perceived correlations in human ability show no correlation in LLMs. Additionally, capabilities of LLMs that can be described by human ability may emerge or dissipate as the model's scale changes.
\section*{Limitations}
The limitation of this paper lies in the small number of ultra-large-scale models. At this stage, there are very few ultra-large Language Models (LLMs) with more than 50 billion parameters, which are insufficient to support ranking-based clustering research. Therefore, it is difficult to obtain stable results when applying the methods proposed in this paper to the study of models with more than 20 billion parameters. Thus, the analysis of the capabilities of ultra-large-scale models constitutes a limitation of this paper.
\section{acknowledge}
The corresponding author is Cheng Daning.  
\bibliography{custom}
%
\appendix

\section{Benchmark Summary\label{sec:appendix}}

 AFQMC\cite{zhang2022fengshenbang} ; A Chinese semantic similarity task that determines whether two sentences are semantically equivalent. ; Text reasoning 
 
        AGIEval\cite{zhong2023agieval} ; The dataset originates from 20 official, open, and high-standard entrance and qualification examinations. ; Examination 
        
        ARC-C\cite{clark2018think} ; A multiple-choice question answer dataset, which includes questions from third to ninth grade science exams. ; Examination 
        
        AX$_g$\cite{wang2019superglue} ; Determine which noun the pronoun refers to based on the given sentence and pronoun. ; Analysis 
        
        BBH\cite{suzgun2022challenging} ; Focused on the 23 challenging tasks of BIG Bench. ; Comprehensive Reasoning 
        
        Common- sense QA\cite{clark2019boolq} ; A dataset for common sense question-answering challenges, aimed at testing common sense knowledge. ; Knowledge
        
        C³\cite{sun2020investigating} ; Chinese Free-form Choice Question Machine Reading Comprehension Dataset ; Understanding
        
        C-Eval\cite{huang2024c} ; A dataset that encompasses evaluations across 52 different disciplines and four levels of difficulty. ; Examination 
        
        CHID\cite{zheng2019chid} ; A Chinese idiom reading comprehension task that requires choosing the correct character to complete an idiom. ; Understanding
        
        CMMLU\cite{li2023cmmlu} ; A benchmark covering 67 topics from foundational subjects to advanced professional levels. ; Examination 
        
        CMNLI\cite{xu2020clue} ; A Chinese natural language inference task that requires judging the logical relationship between two sentences. ; Text reasoning 
        
        NQ ; Understand Wikipedia articles, output a long answer (a paragraph or table) or a short answer ("yes" or "no"). ; Knowledge 
        
        COPA ; A causal inference task that requires selecting the correct causal relationship based on the given premise. ; Text reasoning 
        
        CSL ; Abstract generation, keyword generation, and text classification ; Summarization 
        
        Drop ; A QA dataset for testing comprehensive understanding ability. ; Understanding 
        
        Flores ; Benchmark datasets for evaluating machine translation of low-resource languages ; Multilingual 
        
        GAOKAO-bench ; A dataset based on the Chinese college entrance examination questions. ; Examination 
        
        GSM8K ; A dataset of elementary school math application problems ; Mathematical reasoning 
        
        HellaSwag ; A challenge dataset for evaluating common sense natural language inference. ; Common sense reasoning 
        
        Humaneval ; Collection of programming problems for evaluating the correctness of programming model functionality ; Code 
        
        LAMBADA ; A dataset for evaluating the text understanding ability through a word prediction task. ; Analysis 
        
        LCSTS ; Evaluating the similarity between generated summaries and human summaries using the ROUGE metric. ; Summarization 
        
        MATH \cite{hendrycks2021measuring}; A dataset for measuring the mathematical problem-solving ability of machine learning models. ; Mathematical reasoning
        
        MBPP ; Composed of crowdsourced Python programming problems solvable by about 1000 entry-level programmers. ; Code 
        
        MMLU ; The benchmark text covers 57 topics across various disciplines including STEM, humanities, and social sciences. ; Examination 
        
        TriviaQA ; Process text from various sources such as news articles, encyclopedia entries, and blog posts to answer questions. ; Knowledge
        
        OCNLI ; A native Chinese natural language inference dataset ; Text reasoning
        
        OpenBookQA\cite{mihaylov2018can} ; A question-and-answer dataset simulating an open-book examination.. ; Understanding 
        
        PIQA ; Choose the most reasonable solution based on the given scenario and two possible solutions. ; Common sense reasoning 
        
        RACE ; A large-scale reading comprehension dataset collected from English exams in China. ; Understanding
        
        ReCoRD ; A reading comprehension task that extracts answers from the given news article based on the questions. ; Understanding
        
        SIQA ; Request to choose the most reasonable behavior based on the given scenario and three possible subsequent actions. ; Common sense reasoning
        
        BoolQ ; A dataset of question-and-answer pairs where the questions typically query complex, non-factual information. ; Knowledge 
        
        TyDiQA ; A dataset of question and answer pairs covering 11 different languages. ; Multilingual 
        
        WiC ; A word sense disambiguation task, determining whether a polysemous word has the same meaning in two sentences. ; Analysis 
        
        WSC ; An anaphora resolution task, which requires determining which noun a pronoun refers to based on the context. ; Analysis
        
        XSum ; It requires the use of an abstract modeling method aimed at creating a brief, one-sentence news summary. ; Summarization

\end{document}